\RequirePackage{fix-cm}
\documentclass[twocolumn]{svjour3}          % twocolumn
\smartqed  % flush right qed marks, e.g. at end of proof

\usepackage{cite}
\usepackage[pdftex]{graphicx}
\usepackage{ragged2e}
\usepackage[tight,footnotesize]{subfigure}
\usepackage{rotating}
\usepackage{graphicx}
\usepackage{amsmath,amssymb} % define this before the line numbering.
\usepackage{array}
\usepackage{multirow}
\usepackage{colortbl}
\usepackage{amsfonts}
\usepackage{pifont}
\usepackage{xspace}
\usepackage{etoolbox}
\usepackage{overpic}
\usepackage{color}
\usepackage{microtype}

\usepackage{makecell} 

\usepackage{booktabs} % 用于生成三线表
\usepackage{graphicx} % 如果需要插入图片
\usepackage{amsmath}  % 如果需要数学公式
\usepackage{caption}  % 控制表格和图片标题
\usepackage{cite}
\usepackage{amsmath,amssymb,amsfonts}
\usepackage{amssymb}
\usepackage{color}
\usepackage{mathptmx}
\usepackage{array}
\usepackage{graphicx}

\usepackage{hyperref}
\usepackage{times} % 使用 Times 字体，确保表格和正文一致
\usepackage{graphicx}
\usepackage{cite}
\usepackage{adjustbox} % 调整表格大小
\usepackage{cite}
\usepackage{textcomp}

\usepackage[ruled,vlined]{algorithm2e}
\usepackage{algpseudocode}

\newcommand{\figref}[1]{Fig. \ref{#1}}

\newcommand{\orcid}[1]{\href{https://orcid.org/#1}{\includegraphics[width=10pt]{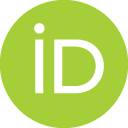}}}

\def\etal{{\em et al}}

\usepackage{hyperref}
\hypersetup{breaklinks=true,citecolor=blue, colorlinks}

\graphicspath{{./Imgs/}}
\DeclareGraphicsExtensions{.pdf,.jpg,.png}

\usepackage{silence}
\journalname{Visual Intelligence}

\begin{document}

\title{Adversarial Attacks on Medical Hyperspectral Imaging Exploiting Spectral Spatial Dependencies and Multiscale Features}

\titlerunning{Short form of title}        % For running head

\author{Yunrui Gu \orcid{0009-0007-7036-9064}        \and
  Zhenzhe Gao \orcid{0009-0008-2545-796X} \and 
  Cong Kong \orcid{0009-0000-7236-1196}\and
  Jiawei Du\orcid{0000-0003-4734-9798}\and
  Zhaoxia Yin\orcid{0000-0003-0387-4806}
}

\authorrunning{Gu \etal} % if too long for running head

\institute{
Yunrui Gu, Zhenzhe Gao, Cong Kong and Zhaoxia Yin are with East China Normal University, China. 
(Email: 51265904060@stu.ecnu.edu.cn, 51255904049@stu.ecnu.edu.cn, 51265904072@stu.ecnu.edu.cn, zxyin@cee.ecnu.edu.cn). \\
Jiawei Du is with Agency for Science, Technology and Research, Singapore. 
(Email: dujiawei@u.nus.edu). \\
Corresponding author: Zhaoxia Yin.
}

\date{Received: date / Accepted: date}

\maketitle

\begin{abstract}
Medical hyperspectral imaging (MHSI) has shown strong potential for disease diagnosis by capturing spectral-spatial information of tissues. While deep learning has substantially improved MHSI classification accuracy, its robustness remains limited due to the well-known trade-off between accuracy and robustness in Deep Neural Networks (DNNs). This issue is particularly critical in MHSI, where reliable prediction depends on local tissue relationships and multiscale spectral-spatial structures. A practical way to improve robustness is to identify the most unstable adversarial examples and incorporate them into adversarial training. However, existing attack methods do not sufficiently exploit these MHSI-specific properties, leading to suboptimal attack effectiveness and limited value for robustness enhancement. To address this gap, we propose a structured adversarial attack framework for MHSI that progressively models its local spectral-spatial dependencies and multiscale hierarchical representations. The proposed method generates anatomically consistent perturbations by modeling neighborhood dependencies and hierarchical spectral-spatial features. Experiments on the brain and choledoch datasets show that our method more effectively degrades lesion-related classification performance in critical tumor regions than existing baselines while maintaining low perturbation magnitude. These results reveal a clinically relevant robustness weakness in current MHSI models and provide stronger adversarial samples for developing targeted defense strategies.
% Please provide 4 to 6 keywords which can be used for indexing purposes.

\end{abstract}

\section{Introduction}

Medical hyperspectral imaging has emerged as a promising technology for clinical diagnosis by capturing rich spectral-spatial information of biological tissues. By integrating both spatial and spectral cues, MHSI enables precise tissue characterization, supporting critical tasks such as tumor detection, vascular visualization, and histopathological segmentation~\cite{lu2014medical,lu2014spectral}. Compared with conventional imaging modalities, MHSI can reveal subtle biochemical and structural variations that are otherwise difficult to observe, making it particularly valuable for early-stage disease detection and fine-grained tissue analysis~\cite{cui2022deep}. With the rapid development of deep learning techniques, data-driven MHSI classification models have achieved substantial improvements in feature representation and diagnostic accuracy~\cite{cui2022deep,tu2024overview}.

Despite these advances, high accuracy alone is insufficient for clinical deployment, where robustness and reliability are equally critical. In practical medical scenarios, incorrect predictions—especially misclassification of lesion regions—may directly affect clinical decision-making, leading to inappropriate treatments or delayed interventions. Recent studies have shown that deep neural networks are inherently vulnerable to adversarial perturbations, where small and imperceptible input changes can cause significant prediction errors~\cite{goodfellow2014explaining,mei2025survey}. Such vulnerabilities are particularly concerning in medical imaging applications, as they undermine diagnostic reliability and may jeopardize patient safety and treatment outcomes~\cite{cheng2024adversarial,baytas2024predicting}. In the context of MHSI, this problem is further amplified due to the complex spectral-spatial representations learned by deep models.

The robustness challenges in MHSI are fundamentally rooted in its unique data characteristics. Unlike natural images or standard hyperspectral remote sensing data, MHSI classification relies heavily on fine-grained local tissue relationships and intricate multiscale spectral-spatial structures. Recent studies have further demonstrated that modeling structured relationships in hyperspectral data, including both local and global dependencies, is essential for robust learning, particularly under challenging conditions such as noisy labels\cite{shi2025learning}. Specifically, the discrimination between pathological and normal tissues often depends on subtle spectral differences embedded within local spatial contexts, where neighboring pixels jointly determine the semantic meaning of tissue regions~\cite{zeng2023microscopic,wei2019medical}. Moreover, clinically relevant patterns span multiple scales, ranging from microscopic cellular structures to macroscopic tissue organization, requiring models to capture hierarchical dependencies across both spatial and spectral dimensions~\cite{lu2014medical,xue2025contemporary}. 

\begin{figure*}[ht]
    \centering
    \includegraphics[width=0.9\textwidth,keepaspectratio]{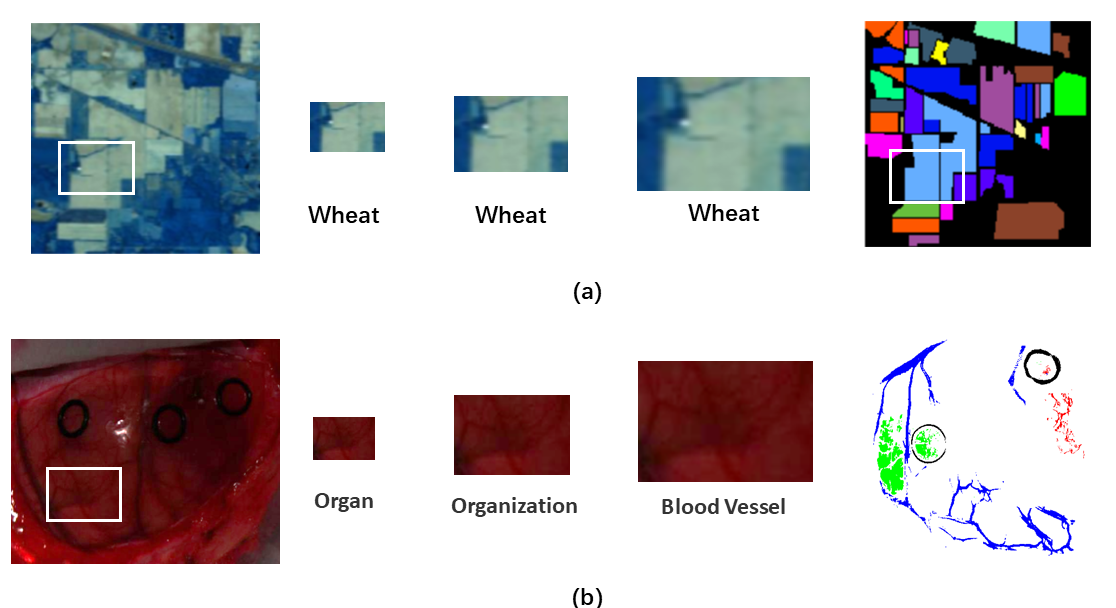}
    %\caption{Comparison of spectral-spatial characteristics between hyperspectral remote sensing and medical hyperspectral imaging.}
    \caption{Comparison of spectral-spatial characteristics between hyperspectral remote sensing and MHSI. Remote sensing images usually exhibit locally homogeneous regions with stable semantics across scales, where image patches of different sampling resolutions correspond to consistent categories (e.g., wheat remains wheat). In contrast, MHSI contains heterogeneous tissue structures, ambiguous local boundaries, and pronounced scale-dependent patterns, where the same region may correspond to different semantic levels (e.g., organ, organization, or blood vessel structures) under different resolutions. These differences suggest that effective adversarial attacks on MHSI should account for both local tissue dependencies and multiscale spectral-spatial representations.}
    \label{intro}
\end{figure*}

As illustrated in Fig.~\ref{intro}, this structural dependency in MHSI is fundamentally different from conventional hyperspectral remote sensing scenarios. In remote sensing images, local regions are typically homogeneous, where pixels within a neighborhood share consistent semantic labels, and classification results remain stable across different spatial scales. Under such assumptions, adversarial perturbations can be effectively modeled at the pixel level. In contrast, MHSI exhibits inherently heterogeneous tissue structures with ambiguous local boundaries, where neighboring pixels may belong to different categories and form mixed compositions. Moreover, the semantic interpretation of tissue regions is strongly scale-dependent: patterns that appear consistent at one scale (e.g., organ-level structures) may correspond to different categories at finer scales (e.g., vascular or cellular structures). This indicates that MHSI classification relies jointly on local structural consistency and multiscale spectral-spatial representations.

%As a result, the vulnerability of MHSI models does not primarily stem from independent pixel-level perturbations, but from the instability of these structured representations under perturbations. This observation suggests that effective adversarial attacks for MHSI should explicitly model both local dependencies and multiscale structures, rather than relying on conventional pixel-wise perturbation strategies.
Therefore, the core vulnerability of MHSI models is not merely that they are sensitive at the pixel level. The deeper problem is that the structured cues underlying lesion recognition are themselves unstable. Whether a spectral variation is clinically meaningful depends on its local tissue neighborhood, while the final diagnostic judgment depends on how such local evidence is assembled across scales. In that sense, an effective attack on MHSI should not be understood simply as a more aggressive pixel-wise optimizer. It should instead operate as a structure-aware mechanism, one that perturbs local dependency patterns and hierarchical spectral-spatial representations in a coordinated way.

Existing research on adversarial attacks in hyperspectral imaging has been largely developed in the context of remote sensing. In these scenarios, deep learning models—particularly convolutional neural networks (CNNs)—have become the dominant approach for hyperspectral classification, effectively learning spectral-spatial features from high-dimensional data~\cite{khan2018modern}. Correspondingly, most adversarial attack methods focus on pixel-level perturbations, assuming that classification decisions are primarily determined by individual spectral signatures. However, such assumptions do not hold in medical hyperspectral imaging, where spatial dependencies between neighboring pixels play a crucial role in preserving anatomical structures and ensuring accurate diagnosis~\cite{khan2021trends,wei2023cat}. As a result, directly applying remote sensing-based adversarial methods to MHSI often fails to capture the intrinsic structural vulnerabilities of medical data, leading to limited attack effectiveness and insufficient insights into model robustness.

Furthermore, MHSI data exhibit richer and more complex spectral-spatial variations compared to remote sensing images. Variations in tissue composition, vascular structures, and pathological regions introduce highly heterogeneous patterns that require multiscale analysis for accurate interpretation. Fine-grained tumor structures are better characterized at local scales, while broader tissue organization and contextual information emerge at larger scales. This inherent multiscale nature further distinguishes MHSI from conventional hyperspectral imaging tasks and highlights the need for attack strategies that explicitly consider hierarchical feature representations rather than isolated perturbations.

Despite these observations, existing adversarial attack methods do not adequately address the unique challenges posed by MHSI. Most approaches are limited to pixel-wise perturbations and fail to exploit the underlying local dependencies and multiscale structures that are critical for medical image analysis~\cite{mfcanet}. This gap motivates the development of more structured and domain-aware adversarial strategies.

In this work, we propose a structured adversarial attack framework specifically tailored for MHSI. The proposed approach progressively models the intrinsic structural properties of MHSI data, first capturing local pixel dependencies to ensure structural consistency, and then extending to multiscale representations to target hierarchical vulnerabilities. By aligning the attack design with the intrinsic characteristics of medical hyperspectral data, our method provides deeper insights into the robustness limitations of MHSI models and offers stronger adversarial examples for developing targeted defense strategies.

In this study, we make the following key contributions:
\begin{itemize}
    %\item We propose a unified adversarial attack framework for MHSI that systematically leverages both local pixel dependencies and multiscale spectral–spatial characteristics.
    \item We identify a previously underexplored robustness issue in MHSI: clinically critical errors are governed by structured lesion representations rather than by independent pixel perturbations alone.  
    \item We formulate a structure-aware adversarial attack framework that explicitly targets the two key factors underlying MHSI prediction, namely local tissue dependency and multiscale spectral-spatial representation.
    %\item We introduce two complementary mechanisms that capture intrinsic properties of medical hyperspectral data, enabling structured and anatomically consistent adversarial perturbations.
    %\item We demonstrate through extensive experiments that our method exposes a clinically critical vulnerability, namely severe lesion misclassification under imperceptible perturbations.
    \item We instantiate this formulation with two coordinated attack components, a local dependency attack and a multiscale attack, which jointly generate coherent perturbations aligned with MHSI-specific decision structure.
    \item We verify that under bounded perturbations, the proposed attack more effectively induces lesion-region misclassification than representative baselines across multiple MHSI classifiers and defense models.
\end{itemize}

\section{Related Works}
\subsection{Hyperspectral Image Classification}
Hyperspectral image (HSI) classification has evolved from traditional machine learning methods, such as Principal Component Analysis (PCA), Support Vector Machines (SVM), and k-Nearest Neighbors (KNN), to deep learning approaches that automatically learn spectral-spatial representations from high-dimensional data~\cite{li2019deep,kumar2020feature}. While traditional methods are computationally efficient, they often rely on handcrafted features and struggle to capture the complex correlations inherent in hyperspectral data.

Deep learning models, particularly convolutional neural networks (CNNs), have significantly improved HSI classification performance by exploiting spectral-spatial dependencies. Early approaches employed 2D-CNNs to extract spatial features, followed by 3D-CNNs that jointly model spectral and spatial information. To balance computational efficiency and representation capacity, hybrid architectures such as HybridSN combine 2D and 3D convolutions, while models like Spectral-Spatial Residual Network (SSRN) further enhance performance through residual learning~\cite{hybridsn2020,ssrn2018}. More recent methods, including Spectral-Spatial Fully Convolutional Networks (SSFCN) and transformer-based architectures, focus on capturing long-range dependencies and multiscale representations in hyperspectral data~\cite{ssfcns2020}.

Despite these advances, existing HSI classification models are primarily developed for remote sensing scenarios, where the focus is on large-scale spatial patterns and global spectral variations. As a result, these models typically emphasize global feature extraction and often treat pixel-level information as relatively independent. Such design assumptions are not well aligned with MHSI, where classification relies on fine-grained local tissue relationships and subtle spectral variations embedded within structured anatomical contexts. Consequently, directly applying these models or their associated assumptions to MHSI may fail to capture the intrinsic structural dependencies that are critical for accurate medical diagnosis and robustness analysis.

\subsection{Medical Hyperspectral Image Classification}

MHSI classification has become an essential component in medical imaging, providing critical support for disease diagnosis and monitoring. Due to the high dimensionality and complex spectral-spatial characteristics of hyperspectral data, a variety of deep learning methods have been developed to effectively extract informative features. Convolutional neural networks (CNNs) have demonstrated strong capability in modeling hierarchical spatial features and have been widely applied in MHSI classification tasks.

For instance, Huang et al.~\cite{Huang2020} proposed a CNN-based framework for blood cell classification that integrates modulated Gabor wavelets to capture multiscale and orientation-specific features, effectively modeling local spatial dependencies in hyperspectral data. More recently, transformer-based methods have been introduced to capture long-range dependencies across spectral bands. Zeng et al.~\cite{zeng2023microscopic} proposed a fusion transformer framework that combines CNNs and transformers in parallel, where the CNN branch extracts spatial features and the transformer branch models spectral relationships, leading to improved performance in MHSI classification.

In addition to model architecture design, explicitly modeling pixel dependencies has been shown to be critical for accurate medical hyperspectral analysis. Xie et al.~\cite{xie2023exploring} proposed a deformable framework for histopathological image segmentation that preserves local spatial relationships, significantly improving the segmentation of cellular structures. Similarly, Wei et al.~\cite{wei2019medical} developed an end-to-end fusion network (EtoE-Fusion) that jointly captures global and local features, enabling more robust classification through multiscale feature integration.

These studies collectively highlight two key characteristics of MHSI data: strong local pixel dependencies and rich multiscale spectral-spatial structures. While these properties are essential for achieving high classification accuracy, they also imply that MHSI models are inherently sensitive to perturbations that disrupt local consistency or multiscale representations. In particular, small but structured changes in local neighborhoods or across scales may lead to significant shifts in model predictions, revealing a unique form of vulnerability that is not adequately captured by conventional pixel-level assumptions. This observation motivates the need for adversarial attack strategies that explicitly account for these intrinsic structural properties of MHSI.

\subsection{Adversarial Attacks on Hyperspectral Images}

Adversarial attacks on hyperspectral images have attracted increasing attention due to their ability to induce severe misclassification through imperceptible perturbations. Most existing methods have been developed in the context of remote sensing, where hyperspectral classification is typically modeled based on spectral signatures or global spatial patterns. As a result, these approaches generally rely on pixel-level perturbations and assume relatively weak dependencies between neighboring pixels.

From a methodological perspective, existing hyperspectral adversarial attacks can be broadly categorized into three groups. The first group focuses on pixel-level perturbations, where adversarial noise is independently applied to individual pixels based on gradient information. For example, Shi et al.~\cite{Shi2022} demonstrated that manipulating spectral bands at the pixel level can effectively degrade classification performance. However, such methods overlook the strong spatial dependencies between neighboring pixels, which are essential for preserving tissue structures in MHSI.

The second group attempts to incorporate global or contextual information into the attack process. Zhao et al.~\cite{mfcanet} proposed a multifeature collaborative adversarial attack that perturbs both spectral and spatial features, while Tu et al.~\cite{rcca} introduced a context-aware framework that adjusts global representations to generate adversarial samples. Although these methods improve attack effectiveness in remote sensing scenarios, they primarily operate at a global level and fail to explicitly model fine-grained local pixel interactions, which are critical for accurate medical hyperspectral analysis.

The third group considers region-level or object-level perturbations to enhance transferability. For instance, Shi et al.~\cite{ssfgsm} proposed a universal object-level adversarial attack targeting local regions, and Zhang et al.~\cite{Zhang2023} introduced a multiloss attack that perturbs multimodal spectral-spatial features. Despite these advances, such approaches still rely on coarse-grained perturbation strategies and do not account for the hierarchical multiscale structures inherent in MHSI data.

Some recent works have explored more advanced attack or defense strategies, such as GAN-based frameworks that generate adversarial samples by modeling spectral-spatial features~\cite{hynu2023patent}. While these approaches partially consider both spectral and spatial characteristics, they are still not specifically designed to capture the structured local dependencies and multiscale representations required in medical hyperspectral imaging.

In summary, existing adversarial attack methods for hyperspectral images are largely designed for remote sensing applications and are primarily based on pixel-level, global, or coarse-grained perturbations. These approaches do not adequately model the strong local pixel dependencies and multiscale spectral-spatial structures that are fundamental to MHSI. Consequently, they fail to fully expose the intrinsic robustness vulnerabilities of MHSI models, highlighting the need for structure-aware adversarial attack frameworks tailored to medical hyperspectral imaging.

\section{Method}
\begin{figure*}[ht]
    \centering
    \includegraphics[width=\textwidth,keepaspectratio]{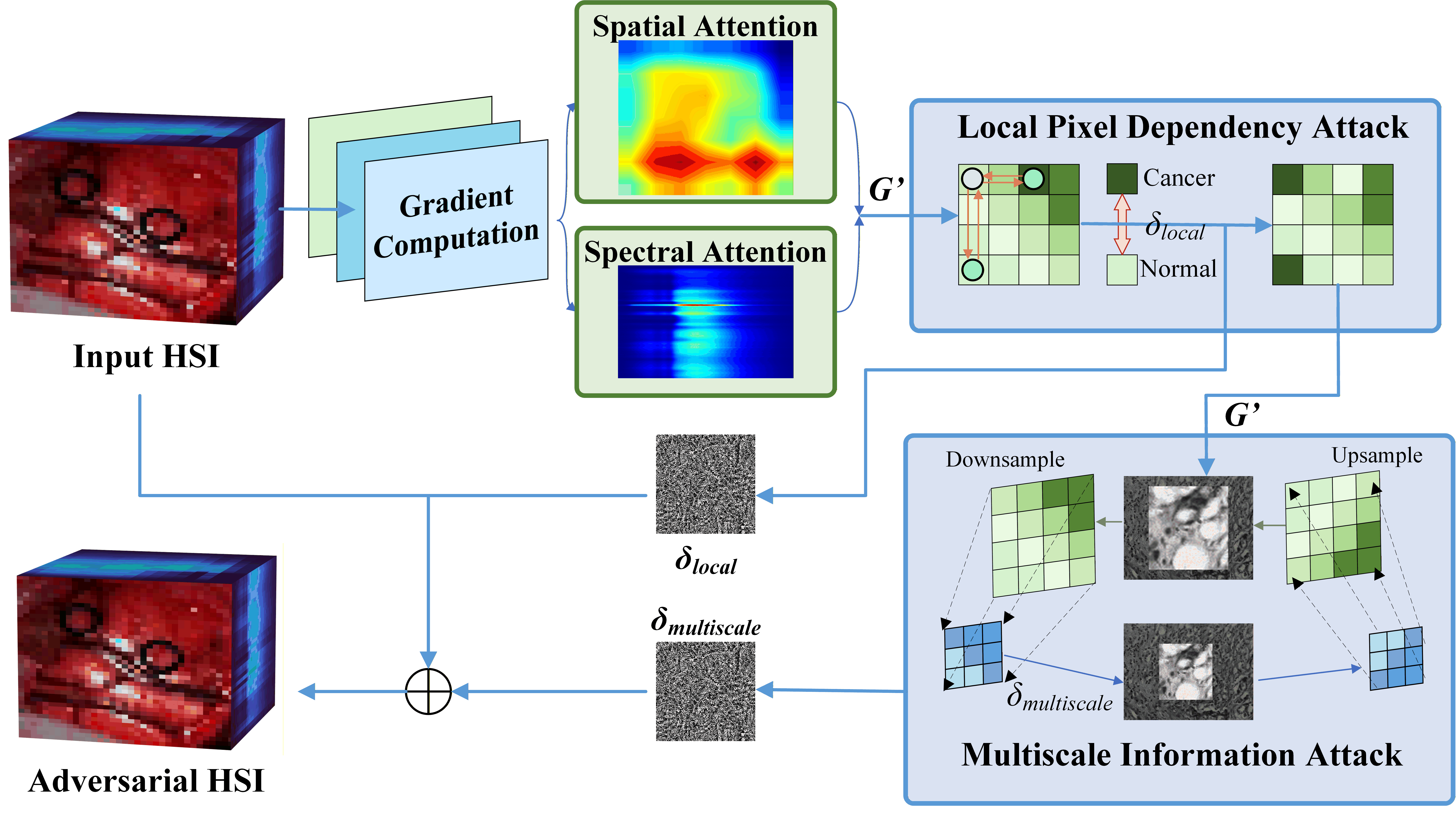}
    %\caption{The proposed adversarial attack framework for MHSI classification.}
    \caption{Overview of the proposed adversarial attack framework for MHSI classification. The method first applies spectral-spatial attention in the gradient space to identify informative regions and spectral bands. It then combines a Local Pixel Dependency Attack, which enforces perturbation coherence within local neighborhoods, with a Multiscale Information Attack, which perturbs representations across different resolutions. By integrating these two components, the framework generates structured adversarial examples that better target the local dependency and hierarchical spectral-spatial properties of MHSI data.}
    \label{fig:framework}
\end{figure*}

In this section, we present an attention-guided adversarial attack framework tailored for MHSI, motivated by its intrinsic spectral-spatial characteristics, including strong local pixel dependencies and hierarchical multiscale representations. In Sec.\ref{3.1}, we first introduce a spectral-spatial attention mechanism that operates in the gradient space to emphasize structurally and spectrally important regions. In Sec.\ref{3.2}, we build upon this design and propose a Local Pixel Dependency Attack, which extends conventional point-wise perturbations to local neighborhoods by aggregating attention-weighted gradients, thereby preserving spatial coherence while effectively misleading the model. In Sec.\ref{3.3}, we further develop a Multiscale Information Attack that generates perturbations across multiple resolutions to disrupt hierarchical spectral-spatial representations learned by MHSI models. In Sec.\ref{3.4}, we integrate these components into a unified adversarial attack framework, where attention-guided local perturbations and multiscale perturbations jointly contribute to the final adversarial example.
\subsection{Spectral-Spatial Attention Mechanism}
\label{3.1}
To better exploit the intrinsic spectral–spatial characteristics of MHSI, we introduce a spectral-spatial attention mechanism that adaptively emphasizes critical spatial regions and informative spectral bands during adversarial perturbation generation.

Unlike conventional attention mechanisms that operate on feature representations, our design directly modulates the gradient space, enabling structure-aware perturbations aligned with the model's decision sensitivity.

Given an input hyperspectral image $x \in \mathbb{R}^{D \times H \times W}$ and its corresponding loss function $\mathcal{L}(x, y)$, we first compute the gradient:

\begin{equation}
G = \nabla_x \mathcal{L}(x, y).
\label{eq:gradient}
\end{equation}

\textbf{Spatial Attention.}
Following the structure of the Convolutional Block Attention Module\cite{cbam}, 
we aggregate spectral information via channel-wise pooling to compute spatial attention:

\begin{equation}
A_{\text{spatial}} = \sigma \left( \text{Conv}_{2D} \left( \text{AvgPool}_D(G) + \text{MaxPool}_D(G) \right) \right),
\label{eq:spatial_attention}
\end{equation}
where $A_{\text{spatial}} \in \mathbb{R}^{H \times W}$ represents the spatial attention map, and $\sigma(\cdot)$ denotes the sigmoid activation function.

\textbf{Spectral Attention.}
To model the importance of spectral bands, we compute:

\begin{equation}
A_{\text{spectral}} = \sigma \left( \text{MLP} \left( \text{GlobalPool}_{H,W}(G) \right) \right),
\label{eq:spectral_attention}
\end{equation}
where $A_{\text{spectral}} \in \mathbb{R}^{D}$ captures the contribution of each spectral channel.

\textbf{Attention-weighted Gradient.}
The final gradient used for adversarial perturbation is obtained by combining both attention mechanisms:

\begin{equation}
G' = A_{\text{spectral}} \odot A_{\text{spatial}} \odot G,
\label{eq:attention_gradient}
\end{equation}
where $\odot$ denotes element-wise multiplication with broadcasting.

The attention-weighted gradient $G'$ is then used in subsequent attack modules, including the Local Pixel Dependency Attack and the Multiscale Information Attack, to guide the generation of structured adversarial perturbations.

\subsection{Local Pixel Dependency Attack}
\label{3.2}

In MHSI, classification decisions rely heavily on local spatial coherence, where neighboring pixels jointly determine tissue semantics. As a result, perturbations that ignore such dependencies may break anatomical structures and become less effective or easily detectable. To address this issue, we propose the Local Pixel Dependency Attack, which explicitly incorporates local spatial relationships into adversarial perturbation generation.

Unlike conventional pixel-wise attacks, our method enforces structural consistency by aggregating gradients within local neighborhoods. Moreover, to further align perturbations with model sensitivity, we employ the attention-weighted gradient $G'$ defined in Equ.~(\ref{eq:attention_gradient}), which integrates both spectral and spatial importance.

Given an input hyperspectral image $\mathbf{x}$ and its ground-truth label $y$, we directly use the attention-weighted gradient $G'$ defined in Equ.~(\ref{eq:attention_gradient}) for subsequent local aggregation.

To preserve local structural consistency, we define a spatial neighborhood $\mathcal{W}(i,j)$ centered at each pixel $(i,j)$. The perturbation direction is obtained by averaging the attention-weighted gradients within this local window:

\begin{equation}
\bar{G}_{i,j} = \frac{1}{N_{i,j}} \sum_{(i', j') \in \mathcal{W}(i,j)} G'(i', j'),
\label{eq:local_avg}
\end{equation}
where $\mathcal{W}(i,j)$ denotes a $K \times K$ square neighborhood centered at pixel $(i,j)$, and $N_{i,j} = K^2$ is the number of pixels within this window.

The adversarial example is then generated by applying the locally aggregated perturbation:

\begin{equation}
\mathbf{x}_{\text{adv}} = \mathbf{x} - \epsilon \cdot \bar{G},
\label{eq:local_attack}
\end{equation}
where $\epsilon$ controls the perturbation magnitude.

To further refine the adversarial perturbation, we adopt an iterative update scheme:

\begin{equation}
\mathbf{x}^{(t+1)} = \mathbf{x}^{(t)} - \epsilon \cdot \bar{G}^{(t)},
\label{eq:local_iter}
\end{equation}
where $\bar{G}^{(t)}$ is computed according to Equ.~(\ref{eq:local_avg}) at iteration $t$.

This local aggregation mechanism ensures that perturbations respect spatial continuity and preserve anatomical structures, making them more imperceptible while remaining highly effective. In addition, the integration of attention-weighted gradients further guides perturbations toward spectrally and spatially informative regions, enhancing attack precision.

Formally, we denote the resulting local perturbation as:
\begin{equation}
\delta_{\text{local}} = \epsilon \cdot \bar{G},
\end{equation}
where $\bar{G}$ represents the locally aggregated gradient defined in Equ.(\ref{eq:local_avg}).

In the targeted attack setting, the objective is to drive the prediction toward a specific target label $\hat{y}$, which is achieved by minimizing the following loss:

\begin{equation}
\mathcal{L}(\mathbf{x}_{\text{adv}}, \hat{y}) = - \log P(\hat{y} \mid \mathbf{x}_{\text{adv}}),
\label{eq:target_loss}
\end{equation}
where the optimization is still guided by the attention-weighted gradient $G'$.

\subsection{Multiscale Information Attack}  
\label{3.3}

While modeling local pixel dependencies improves the structural consistency of adversarial perturbations, MHSI models also rely heavily on hierarchical spectral-spatial representations across multiple scales. Discriminative patterns span from fine-grained local structures to coarse tissue-level organization, making multiscale representations a critical factor in model decision-making. 

To exploit this property, we extend the attack to a multiscale setting and propose a Multiscale Information Attack, which generates perturbations across multiple resolutions to effectively target these hierarchical representations.

Given an input hyperspectral image $\mathbf{x} \in \mathbb{R}^{B \times D \times H \times W}$, we utilize the attention-weighted gradient $G'$ defined in Equ.~(\ref{eq:attention_gradient}) to guide perturbation generation across different scales.

For each scale factor $s \in \mathcal{S}$, the input is first downsampled:

\begin{equation}
\mathbf{x}_{\text{down}}^{(s)} = \text{Downsample}(\mathbf{x}, s).
\label{eq:downsample}
\end{equation}

To ensure consistency with the original optimization objective, we compute the attention-weighted gradient at the original resolution and propagate it to the scale space:

\begin{equation}
{G'}_{\text{down}}^{(s)} = \text{Downsample}(G', s).
\label{eq:grad_down}
\end{equation}

Perturbations are then introduced in the scale space as:

\begin{equation}
\mathbf{x}_{\text{pert}}^{(s)} = \mathbf{x}_{\text{down}}^{(s)} - \epsilon \cdot {G'}_{\text{down}}^{(s)}.
\label{eq:perturb_scale}
\end{equation}

After perturbation, the image is upsampled back to the original resolution:

\begin{equation}
\mathbf{x}_{\text{up}}^{(s)} = \text{Upsample}(\mathbf{x}_{\text{pert}}^{(s)}, (H, W)).
\label{eq:upsample}
\end{equation}

The perturbation at each scale is computed as:

\begin{equation}
\mathbf{p}_s = \mathbf{x}_{\text{up}}^{(s)} - \mathbf{x}.
\label{eq:ps}
\end{equation}

To incorporate multiscale information, perturbations across all scales are aggregated:

\begin{equation}
\mathbf{p} = \sum_{s \in \mathcal{S}} \mathbf{p}_s.
\label{eq:multi_sum}
\end{equation}

Finally, the adversarial example is generated as:

\begin{equation}
\mathbf{x}_{\text{adv}} = \mathbf{x} - \mathbf{p}.
\label{eq:multi_final}
\end{equation}

By introducing perturbations at multiple resolutions and aligning them with attention-weighted gradients, the Multiscale Information Attack effectively disrupts both fine-grained and coarse-grained spectral-spatial features. This hierarchical perturbation strategy is particularly suited for medical hyperspectral data, where diagnostic cues are distributed across multiple spatial and spectral scales.
\subsection{Adversarial Attack Framework}
\label{3.4}

As illustrated in \figref{fig:framework}, the proposed adversarial attack framework follows a progressive design that explicitly models the structural properties of MHSI data. Specifically, we first generate a locally consistent perturbation by modeling pixel dependencies, guided by the attention-weighted gradient defined in Sec.\ref{3.1}. This step preserves anatomical structure while effectively misleading the model. 

Based on this locally structured perturbation, we further extend the attack to a multiscale setting, enabling the perturbation to capture hierarchical spectral-spatial representations across different resolutions.

Formally, the final adversarial perturbation is constructed by integrating the local structural perturbation and its multiscale extensions:

\begin{equation}
\delta_{\text{final}} = \delta_{\text{local}} + \delta_{\text{multiscale}},
\label{eq:final_delta}
\end{equation}
where $\delta_{\text{multiscale}}$ is generated based on attention-guided perturbations across multiple scales.

The adversarial example is then obtained as:

\begin{equation}
\mathbf{x}_{\text{adv}} = \mathbf{x} + \delta_{\text{final}},
\label{eq:final_adv}
\end{equation}
where $x$ denotes the original hyperspectral image and $x_{\text{adv}}$ is the corresponding adversarial example.

The general process of the proposed attack method can be briefly summarized as Algorithm~\ref{algo}:
% 无需引入 algpseudocode/algorithm2e 等包，仅需基础 LaTeX 环境
\begin{algorithm}[t]
\caption{Attention-Guided Adversarial Attack Framework for MHSI}
\label{algo}

\KwIn{HSI data $\mathbf{x}$, target $\hat{y}$, model $f$, perturbation budget $\epsilon$, iterations $T$, scales $\mathcal{S}$}
\KwOut{Adversarial sample $\mathbf{x}_{adv}$}

$\mathbf{x}_{adv} \gets \mathbf{x}$\;

\For{$t \gets 1$ \KwTo $T$}{

    \tcp{Compute attention-weighted gradient}
    $G \gets \nabla_{\mathbf{x}} \mathcal{L}(\mathbf{x}_{adv}, \hat{y})$\;
    $G' \gets A_{\text{spectral}} \odot A_{\text{spatial}} \odot G$\;

    \tcp{Local Pixel Dependency Attack}
    \ForEach{pixel $(i,j)$ in $\mathbf{x}_{adv}$}{
        $\mathcal{W}(i,j) \gets$ local window\;
        $\bar{G}_{i,j} \gets \frac{1}{N_{i,j}} \sum_{(i',j') \in \mathcal{W}(i,j)} G'(i',j')$\;
        $\mathbf{x}_{adv}(i,j) \gets \mathbf{x}_{adv}(i,j) - \epsilon \bar{G}_{i,j}$\;
    }

    \tcp{Multiscale Information Attack}
    $\mathbf{p} \gets 0$\;

    \ForEach{$s \in \mathcal{S}$}{
        $\mathbf{x}_{down} \gets \mathrm{Downsample}(\mathbf{x}_{adv}, s)$\;
        $G'_{down} \gets \mathrm{Downsample}(G', s)$\;
        $\mathbf{x}_{pert} \gets \mathbf{x}_{down} - \epsilon G'_{down}$\;
        $\mathbf{x}_{up} \gets \mathrm{Upsample}(\mathbf{x}_{pert})$\;
        $\mathbf{p} \gets \mathbf{p} + (\mathbf{x}_{up} - \mathbf{x}_{adv})$\;
    }

    $\mathbf{x}_{adv} \gets \mathbf{x}_{adv} + \mathbf{p}$\;
}

\Return{$\mathbf{x}_{adv}$}\;
\end{algorithm}

\section{Experiments}
\subsection{Datasets}
\subsubsection{In-Vivo Hyperspectral Human Brain Image Dataset}
The In-Vivo Hyperspectral Human Brain Image Database for Brain Cancer Detection consists of 36 hyperspectral images collected from 22 neurosurgical operations\cite{fabelo2019vivo}. It covers four annotated classes: normal tissue, tumor tissue, blood vessels, and background elements. The images span the visual and near-infrared spectrum from 400 to 1000 nm, providing over 300,000 labeled spectral signatures. Labels were generated using a semi-automatic methodology based on the Spectral Angle Mapper (SAM) algorithm, cross-referenced with histopathological evaluations. This dataset serves as a significant resource for developing machine learning models for brain tumor classification and guiding real-time surgical decisions.

\subsubsection{Multidimensional Choledoch (MDC) Dataset}
Multidimensional Choledoch (MDC) Dataset includes 880 hyperspectral scenes collected from 174 individuals, comprising 689 scenes with partial cancer regions (L), 49 with complete cancerous areas (N), and 142 without cancer (P)\cite{zhang2019multidimensional}. This dataset only uses binary classification to determine the cancer region from the normal region. The hyperspectral data were captured using a system with a 20× objective lens, covering wavelengths from 450 nm to 1000 nm with 60 spectral bands per scene. Each hyperspectral image was resized to 256×320 pixels to enhance computational efficiency. 
\subsection{Experimental Setup}
In this study, we performed experiments on hyperspectral image datasets, specifically targeting medical image classification tasks. To reduce the data dimensionality and extract the most important spectral features, Principal Component Analysis (PCA) was applied, reducing the spectral dimensions to 20 components. This reduction in dimensionality helps minimize computational overhead while retaining the essential spectral information for classification.

We adopt a patch-based classification setting. Specifically, each hyperspectral image is decomposed into overlapping patches using a sliding window of size 11×11, and each patch is assigned the label of its center pixel. In this way, the task is formulated as pixel-wise classification with patch-level inputs, allowing the model to capture local spectral–spatial features effectively. Zero-padding is applied at the image borders to handle edge pixels that lack sufficient neighboring context.

For dataset splitting, we randomly divide the samples into training and testing sets with a ratio of 80\% and 20\%, respectively. We preserve the class distribution in each split to ensure a balanced evaluation, and no patch overlap across train/test images.

For adversarial attacks, all methods are evaluated under the same attack setting to ensure fair comparison. We set the perturbation budget to $\epsilon$ = 0.01 and the number of iterations to T = 20, where $\epsilon$ denotes the maximum perturbation bound under the $L_{\infty}$ norm. The relatively small $\epsilon$ is chosen considering the high sensitivity of medical hyperspectral data, where even minor perturbations may lead to significant clinical implications. This setting ensures that the generated adversarial perturbations remain highly imperceptible while still being effective.

\begin{table*}[ht]
\centering
%\caption{Performance Comparison of Adversarial Attacks on the Brain Dataset Across Different Models (mean $\pm$ std over five runs).}
\caption{Comparison of adversarial attack performance on the Brain dataset across different target classifiers and defense models (mean ± std over five runs). Lower tumor accuracy indicates a stronger attack on lesion-related regions. Across most models, the proposed method achieves the largest reduction in tumor-region accuracy while preserving relatively high performance on non-lesion classes, suggesting that it more effectively targets lesion-sensitive decision cues in MHSI.}
\resizebox{\textwidth}{!}{

\begin{tabular}{lcccccccccc}
\toprule
Target Model & Attack Method & \makecell{Normal\\Tissue(↑)} & \makecell{Tumor\\Tissue(↓)} & \makecell{Hyper\\vascularized(↑)} & \makecell{Background\\(↑)} & OA(↑) & AA(↑) & KAPPA(↑) & L0(↓) & L2(↓) \\
\midrule

\multirow{4}{*}{HybridSN\cite{hybridsn2020}}
& MfcaNet\cite{mfcanet} & 100 & 14.19$\pm$0.52 & 100 & 100 & 96.61$\pm$0.08 & 78.45$\pm$0.21 & 95.02$\pm$0.09 & 3428 & 12.2 \\
& SSA\cite{hynu2023patent} & 100 & 13.47$\pm$0.47 & 100 & 100 & 96.61$\pm$0.07 & 78.42$\pm$0.25 & 95.01$\pm$0.11 & 4687 & 13.6 \\
& SS-FGSM\cite{ssfgsm} & 100 & 17.84$\pm$0.61 & 100 & 100 & 96.61$\pm$0.06 & 78.44$\pm$0.22 & 95.02$\pm$0.10 & 3274 & 11.3 \\
& Ours & 100 & \textbf{7.98$\pm$0.41} & 100 & 100 & 96.59$\pm$0.09 & 78.30$\pm$0.24 & 94.98$\pm$0.12 & \textbf{1896} & \textbf{7.8} \\

\midrule\addlinespace[0.5em]

\multirow{4}{*}{SSRN\cite{ssrn2018}}
& MfcaNet\cite{mfcanet} & 100 & 14.75$\pm$0.49 & 100 & 100 & 96.61$\pm$0.07 & 78.47$\pm$0.23 & 95.03$\pm$0.08 & 3512 & 12.5 \\
& SSA\cite{hynu2023patent} & 100 & 21.86$\pm$0.58 & 100 & 100 & 96.63$\pm$0.08 & 78.68$\pm$0.27 & 95.08$\pm$0.11 & 4825 & 14.1 \\
& SS-FGSM\cite{ssfgsm} & 100 & 11.94$\pm$0.46 & 100 & 100 & 96.60$\pm$0.06 & 78.32$\pm$0.21 & 95.03$\pm$0.10 & 3341 & 11.5 \\
& Ours & 100 & \textbf{4.65$\pm$0.38} & 100 & 100 & 96.58$\pm$0.09 & 78.90$\pm$0.26 & 94.95$\pm$0.13 & \textbf{1958} & \textbf{8.1} \\

\midrule\addlinespace[0.5em]

\multirow{4}{*}{SACNet\cite{sacnet}}
& MfcaNet\cite{mfcanet} & 100 & 18.28$\pm$0.54 & 100 & 100 & 96.62$\pm$0.07 & 78.57$\pm$0.24 & 95.05$\pm$0.09 & 3395 & 11.9 \\
& SSA\cite{hynu2023patent} & 100 & 25.67$\pm$0.63 & 100 & 100 & 96.64$\pm$0.08 & 78.82$\pm$0.28 & 95.12$\pm$0.12 & 4973 & 14.7 \\
& SS-FGSM\cite{ssfgsm} & 100 & 16.54$\pm$0.57 & 100 & 100 & 96.62$\pm$0.06 & 78.51$\pm$0.23 & 95.04$\pm$0.10 & 3189 & 11 \\
& Ours & 100 & \textbf{8.46$\pm$0.43} & 100 & 100 & 96.59$\pm$0.09 & 78.32$\pm$0.25 & 94.99$\pm$0.12 & \textbf{2027} & \textbf{8.5} \\

\midrule\addlinespace[0.5em]

\multirow{4}{*}{UAGCN\cite{iclrhyp}}
& MfcaNet\cite{mfcanet} & 100 & 15.08$\pm$0.51 & 100 & 100 & 96.68$\pm$0.07 & 78.61$\pm$0.22 & 95.07$\pm$0.09 & 3315 & 11.8 \\
& SSA\cite{hynu2023patent} & 100 & 20.15$\pm$0.59 & 100 & 100 & 96.71$\pm$0.08 & 78.96$\pm$0.27 & 95.15$\pm$0.11 & 4728 & 14.3 \\
& SS-FGSM\cite{ssfgsm} & 100 & 14.26$\pm$0.55 & 100 & 100 & 96.66$\pm$0.06 & 78.55$\pm$0.23 & 95.02$\pm$0.10 & 3092 & 10.9 \\
& Ours & 100 & \textbf{6.72$\pm$0.40} & 100 & 100 & 96.62$\pm$0.09 & 78.41$\pm$0.24 & 94.97$\pm$0.13 & \textbf{1975} & \textbf{7.9} \\

\midrule\addlinespace[0.5em]

\multirow{4}{*}{Dual-Stream\cite{dual-stream}}
& MfcaNet\cite{mfcanet} & 100 & 28.59$\pm$0.66 & 100 & 100 & 96.68$\pm$0.08 & 82.15$\pm$0.30 & 94.22$\pm$0.14 & 3607 & 13 \\
& SSA\cite{hynu2023patent} & 100 & 32.34$\pm$0.72 & 100 & 100 & 95.80$\pm$0.09 & 83.09$\pm$0.31 & 93.75$\pm$0.16 & 4896 & 14.4 \\
& SS-FGSM\cite{ssfgsm} & 100 & 37.45$\pm$0.75 & 100 & 100 & 95.96$\pm$0.08 & 84.36$\pm$0.29 & 94.46$\pm$0.15 & 3425 & 11.8 \\
& Ours & 100 & \textbf{19.25$\pm$0.53} & 100 & 100 & 95.48$\pm$0.10 & 79.81$\pm$0.28 & 92.48$\pm$0.17 & \textbf{2144} & \textbf{8.7} \\

\midrule\addlinespace[0.5em]

\multirow{4}{*}{RCCA\cite{rcca}}
& MfcaNet\cite{mfcanet} & 99.55$\pm$0.08 & 83.61$\pm$0.41 & 99.83$\pm$0.05 & 99.94$\pm$0.03 & 99.09$\pm$0.04 & 95.54$\pm$0.21 & 98.55$\pm$0.05 & 2968 & 10.4 \\
& SSA\cite{hynu2023patent} & 98.27$\pm$0.12 & 75.82$\pm$0.53 & 99.61$\pm$0.07 & 99.37$\pm$0.09 & 98.31$\pm$0.05 & 93.27$\pm$0.28 & 96.91$\pm$0.08 & 4087 & 11.9 \\
& SS-FGSM\cite{ssfgsm} & 99.58$\pm$0.07 & 79.14$\pm$0.49 & 99.89$\pm$0.04 & 99.55$\pm$0.06 & 98.72$\pm$0.05 & 94.68$\pm$0.24 & 97.08$\pm$0.07 & 2854 & 10.9 \\
& Ours & 98.78$\pm$0.10 & \textbf{68.59$\pm$0.38} & 99.31$\pm$0.09 & 98.75$\pm$0.12 & 97.94$\pm$0.06 & 92.95$\pm$0.26 & 95.28$\pm$0.09 & \textbf{1718} & \textbf{7} \\

\midrule\addlinespace[0.5em]

\multirow{4}{*}{WFSS\cite{tang2024wfss}}
& MfcaNet\cite{mfcanet} & 98.83$\pm$0.11 & 71.98$\pm$0.52 & 99.42$\pm$0.08 & 99.58$\pm$0.06 & 98.07$\pm$0.05 & 93.45$\pm$0.23 & 98.04$\pm$0.06 & 2897 & 10.7 \\
& SSA\cite{hynu2023patent} & 99.12$\pm$0.09 & 82.04$\pm$0.44 & 99.97$\pm$0.02 & 99.91$\pm$0.03 & 98.93$\pm$0.04 & 95.29$\pm$0.22 & 98.97$\pm$0.05 & 4149 & 12.2 \\
& SS-FGSM\cite{ssfgsm} & 98.72$\pm$0.12 & 67.95$\pm$0.56 & 99.15$\pm$0.10 & 98.85$\pm$0.11 & 97.86$\pm$0.05 & 93.16$\pm$0.25 & 95.13$\pm$0.08 & 2766 & 10.8 \\
& Ours & 98.72$\pm$0.11 & \textbf{62.32$\pm$0.43} & 98.63$\pm$0.13 & 98.37$\pm$0.15 & 96.91$\pm$0.06 & 90.76$\pm$0.27 & 93.57$\pm$0.10 & \textbf{1684} & \textbf{6.8} \\

\midrule\addlinespace[0.5em]

\multirow{4}{*}{AIAF\cite{aiaf}}
& MfcaNet\cite{mfcanet} & 99.70$\pm$0.05 & 85.20$\pm$0.36 & 99.96$\pm$0.02 & 99.92$\pm$0.03 & 98.72$\pm$0.04 & 96.22$\pm$0.20 & 98.25$\pm$0.05 & 3021 & 10.9 \\
& SSA\cite{hynu2023patent} & 98.90$\pm$0.09 & 81.45$\pm$0.41 & 99.63$\pm$0.06 & 99.71$\pm$0.05 & 98.39$\pm$0.05 & 94.94$\pm$0.23 & 97.62$\pm$0.07 & 4012 & 11.7 \\
& SS-FGSM\cite{ssfgsm} & 99.62$\pm$0.06 & 83.12$\pm$0.39 & 99.80$\pm$0.04 & 99.88$\pm$0.03 & 98.63$\pm$0.04 & 95.64$\pm$0.21 & 98.02$\pm$0.06 & 2879 & 11 \\
& Ours & 99.15$\pm$0.08 & \textbf{70.34$\pm$0.35} & 98.97$\pm$0.11 & 99.21$\pm$0.09 & 98.01$\pm$0.05 & 91.87$\pm$0.24 & 96.62$\pm$0.08 & \textbf{1741} & \textbf{7.1} \\

\midrule\addlinespace[0.5em]

\multirow{4}{*}{S\textsuperscript{3}ANet\cite{s3anet}}
& MfcaNet\cite{mfcanet} & 99.74$\pm$0.05 & 86.02$\pm$0.34 & 99.91$\pm$0.03 & 99.98$\pm$0.02 & 98.77$\pm$0.04 & 96.34$\pm$0.19 & 98.31$\pm$0.05 & 2954 & 10.6 \\
& SSA\cite{hynu2023patent} & 99.20$\pm$0.08 & 82.57$\pm$0.38 & 99.67$\pm$0.05 & 99.61$\pm$0.06 & 98.46$\pm$0.05 & 95.43$\pm$0.22 & 97.73$\pm$0.07 & 4193 & 12.1 \\
& SS-FGSM\cite{ssfgsm} & 99.65$\pm$0.06 & 84.30$\pm$0.36 & 99.84$\pm$0.04 & 99.76$\pm$0.05 & 98.66$\pm$0.04 & 95.96$\pm$0.21 & 98.08$\pm$0.06 & 2798 & 10.7 \\
& Ours & 99.18$\pm$0.07 & \textbf{72.11$\pm$0.33} & 99.13$\pm$0.09 & 98.91$\pm$0.11 & 98.10$\pm$0.05 & 92.81$\pm$0.23 & 96.80$\pm$0.08 & \textbf{1697} & \textbf{6.9} \\

\bottomrule
\end{tabular}
}

\label{gu.t1}
\end{table*}

\begin{table*}[h]
\centering
%\caption{Performance Comparison of Adversarial Attacks on the MDC Dataset Across Different Models (mean $\pm$ std over five runs).}
\caption{Comparison of adversarial attack performance on the MDC dataset across different target classifiers and defense models (mean ± std over five runs). Lower cancer accuracy indicates stronger attack effectiveness on lesion-related regions. The proposed method consistently produces the largest reduction in cancer-region accuracy while preserving high normal-class accuracy and comparable global performance, suggesting that it more effectively disrupts lesion-sensitive decision patterns in MHSI.
}
\resizebox{\textwidth}{!}{
  \begin{tabular}{lcccccccc}
    \toprule
    Target Model & Attack Method & \makecell{Normal\\(↑)} & \makecell{Cancer\\(↓)} & OA(↑) & AA(↑) & KAPPA(↑) & L0(↓) & L2(↓) \\
    \midrule
    \multirow{4}{*}{HybridSN\cite{hybridsn2020}} 
    & MfcaNet\cite{mfcanet} & 100 & 15.79$\pm$0.56 & 87.11$\pm$0.10 & 57.90$\pm$0.28 & 69.47$\pm$0.15 & 1428 & 13.5 \\
    & SSA\cite{hynu2023patent} & 100 & 23.17$\pm$0.63 & 88.29$\pm$0.09 & 61.59$\pm$0.31 & 73.91$\pm$0.17 & 1764 & 13.9 \\
    & SS-FGSM\cite{ssfgsm} & 100 & 18.46$\pm$0.58 & 87.69$\pm$0.10 & 59.23$\pm$0.29 & 71.08$\pm$0.16 & 1289 & 11.7 \\
    & Ours & 100 & \textbf{10.31$\pm$0.42} & 86.55$\pm$0.11 & 55.16$\pm$0.27 & 66.47$\pm$0.18 & \textbf{752} & \textbf{7.8} \\

    \midrule
    \addlinespace[0.5em]

    \multirow{4}{*}{SSRN\cite{ssrn2018}} 
    & MfcaNet\cite{mfcanet} & 100 & 16.54$\pm$0.53 & 87.38$\pm$0.10 & 58.27$\pm$0.26 & 69.96$\pm$0.14 & 1387 & 12.8 \\
    & SSA\cite{hynu2023patent} & 100 & 27.28$\pm$0.67 & 89.09$\pm$0.09 & 63.64$\pm$0.33 & 76.36$\pm$0.18 & 1721 & 13.4 \\
    & SS-FGSM\cite{ssfgsm} & 100 & 28.57$\pm$0.69 & 89.29$\pm$0.08 & 64.29$\pm$0.32 & 77.14$\pm$0.17 & 1244 & 11.2 \\
    & Ours & 100 & \textbf{13.62$\pm$0.45} & 87.72$\pm$0.10 & 56.81$\pm$0.28 & 68.17$\pm$0.16 & \textbf{713} & \textbf{7.4} \\

    \midrule
    \addlinespace[0.5em]

    \multirow{4}{*}{SACNet\cite{sacnet}} 
    & MfcaNet\cite{mfcanet} & 100 & 15.49$\pm$0.55 & 87.09$\pm$0.10 & 57.75$\pm$0.27 & 69.30$\pm$0.15 & 1472 & 13.9 \\
    & SSA\cite{hynu2023patent} & 100 & 19.43$\pm$0.60 & 87.72$\pm$0.09 & 59.72$\pm$0.30 & 71.65$\pm$0.16 & 1802 & 14.3 \\
    & SS-FGSM\cite{ssfgsm} & 100 & 13.81$\pm$0.49 & 86.91$\pm$0.10 & 56.91$\pm$0.28 & 68.27$\pm$0.15 & 1326 & 12 \\
    & Ours & 100 & \textbf{9.56$\pm$0.40} & 86.39$\pm$0.11 & 54.78$\pm$0.26 & 65.74$\pm$0.17 & \textbf{794} & \textbf{8.2} \\

    \midrule
    \addlinespace[0.5em]

    \multirow{4}{*}{UAGCN\cite{iclrhyp}} 
    & MfcaNet\cite{mfcanet} & 100 & 14.88$\pm$0.54 & 87.15$\pm$0.10 & 57.82$\pm$0.27 & 69.35$\pm$0.15 & 1421 & 13.6 \\
    & SSA\cite{hynu2023patent} & 100 & 18.97$\pm$0.59 & 87.79$\pm$0.09 & 60.11$\pm$0.29 & 71.80$\pm$0.16 & 1768 & 14.2 \\
    & SS-FGSM\cite{ssfgsm} & 100 & 13.28$\pm$0.47 & 86.98$\pm$0.10 & 56.85$\pm$0.27 & 68.40$\pm$0.15 & 1298 & 11.5 \\
    & Ours & 100 & \textbf{8.77$\pm$0.39} & 86.48$\pm$0.11 & 54.92$\pm$0.26 & 66.10$\pm$0.17 & \textbf{768} & \textbf{7.9} \\

    \midrule
    \addlinespace[0.5em]

    \multirow{4}{*}{Dual-Stream\cite{dual-stream}} 
    & MfcaNet\cite{mfcanet} & 100 & 42.15$\pm$0.72 & 92.86$\pm$0.08 & 71.08$\pm$0.34 & 85.75$\pm$0.18 & 1398 & 13.1 \\
    & SSA\cite{hynu2023patent} & 100 & 37.87$\pm$0.68 & 92.00$\pm$0.09 & 68.94$\pm$0.33 & 83.25$\pm$0.19 & 1750 & 13.6 \\
    & SS-FGSM\cite{ssfgsm} & 100 & 45.94$\pm$0.75 & 93.06$\pm$0.08 & 72.97$\pm$0.35 & 86.29$\pm$0.17 & 1279 & 11.6 \\
    & Ours & 100 & \textbf{32.64$\pm$0.57} & 91.64$\pm$0.10 & 66.32$\pm$0.31 & 82.35$\pm$0.20 & \textbf{741} & \textbf{7.6} \\

    \midrule
    \addlinespace[0.5em]

    \multirow{4}{*}{RCCA\cite{rcca}} 
    & MfcaNet\cite{mfcanet} & 99.60$\pm$0.06 & 52.71$\pm$0.51 & 92.68$\pm$0.08 & 71.36$\pm$0.29 & 85.03$\pm$0.16 & 1194 & 11 \\
    & SSA\cite{hynu2023patent} & 99.33$\pm$0.08 & 48.52$\pm$0.48 & 92.00$\pm$0.09 & 68.91$\pm$0.31 & 83.20$\pm$0.17 & 1532 & 11.2 \\
    & SS-FGSM\cite{ssfgsm} & 99.70$\pm$0.05 & 56.37$\pm$0.54 & 93.09$\pm$0.08 & 72.69$\pm$0.28 & 86.17$\pm$0.15 & 1093 & 10.3 \\
    & Ours & 98.91$\pm$0.10 & \textbf{44.68$\pm$0.41} & 91.23$\pm$0.10 & 66.34$\pm$0.30 & 81.77$\pm$0.18 & \textbf{624} & \textbf{6.9} \\

    \midrule
    \addlinespace[0.5em]

    \multirow{4}{*}{WFSS\cite{tang2024wfss}} 
    & MfcaNet\cite{mfcanet} & 99.52$\pm$0.07 & 53.02$\pm$0.49 & 92.76$\pm$0.08 & 71.51$\pm$0.28 & 85.16$\pm$0.16 & 1181 & 10.8 \\
    & SSA\cite{hynu2023patent} & 99.18$\pm$0.08 & 49.14$\pm$0.46 & 92.12$\pm$0.09 & 69.07$\pm$0.30 & 83.36$\pm$0.17 & 1519 & 11.1 \\
    & SS-FGSM\cite{ssfgsm} & 99.68$\pm$0.05 & 57.11$\pm$0.52 & 93.17$\pm$0.08 & 73.56$\pm$0.27 & 86.62$\pm$0.15 & 1085 & 10.1 \\
    & Ours & 98.86$\pm$0.11 & \textbf{45.27$\pm$0.39} & 91.33$\pm$0.10 & 66.64$\pm$0.29 & 82.11$\pm$0.18 & \textbf{611} & \textbf{6.7} \\

    \midrule
    \addlinespace[0.5em]

    \multirow{4}{*}{AIAF\cite{aiaf}} 
    & MfcaNet\cite{mfcanet} & 99.50$\pm$0.07 & 74.50$\pm$0.39 & 94.88$\pm$0.06 & 86.75$\pm$0.24 & 92.18$\pm$0.13 & 1042 & 9.7 \\
    & SSA\cite{hynu2023patent} & 99.14$\pm$0.09 & 70.30$\pm$0.42 & 94.21$\pm$0.07 & 84.72$\pm$0.25 & 90.95$\pm$0.14 & 1379 & 10 \\
    & SS-FGSM\cite{ssfgsm} & 99.66$\pm$0.05 & 78.25$\pm$0.37 & 95.25$\pm$0.06 & 88.88$\pm$0.23 & 93.17$\pm$0.12 & 987 & 9.1 \\
    & Ours & 98.79$\pm$0.10 & \textbf{66.12$\pm$0.34} & 93.64$\pm$0.08 & 82.46$\pm$0.26 & 89.15$\pm$0.15 & \textbf{552} & \textbf{6.2} \\

    \midrule
    \addlinespace[0.5em]

    \multirow{4}{*}{S\textsuperscript{3}ANet\cite{s3anet}} 
    & MfcaNet\cite{mfcanet} & 99.42$\pm$0.07 & 75.03$\pm$0.38 & 94.93$\pm$0.06 & 87.02$\pm$0.23 & 92.39$\pm$0.13 & 1026 & 9.5 \\
    & SSA\cite{hynu2023patent} & 99.09$\pm$0.09 & 71.07$\pm$0.41 & 94.27$\pm$0.07 & 85.04$\pm$0.24 & 91.04$\pm$0.14 & 1364 & 9.8 \\
    & SS-FGSM\cite{ssfgsm} & 99.64$\pm$0.05 & 79.06$\pm$0.36 & 95.32$\pm$0.06 & 89.53$\pm$0.22 & 93.44$\pm$0.12 & 972 & 9 \\
    & Ours & 98.85$\pm$0.10 & \textbf{67.26$\pm$0.33} & 93.96$\pm$0.08 & 83.13$\pm$0.25 & 89.89$\pm$0.15 & \textbf{543} & \textbf{6.1} \\

    \bottomrule
  \end{tabular}
}
\label{gu.t2}
\end{table*}

\subsection{Evaluation Metrics}
Since we are conducting medical image adversarial attacks, attacking the lesion area to misclassify it into normal areas will cause the greatest harm to patients and the medical system. Therefore, our evaluation metrics mainly focus on the classification success rate of the lesion areas in each dataset. The lower the classification accuracy on lesion regions, the stronger the effectiveness of the attack. At the same time, we also adopt three commonly used metrics for comprehensive evaluation:

\textbf{Overall Accuracy (OA)} measures the overall proportion of correctly classified pixels. It is defined as:
\begin{equation}
\text{OA} = \frac{\sum_{i=1}^{C} N_{ii}}{\sum_{i=1}^{C} \sum_{j=1}^{C} N_{ij}},
\end{equation}
where \( N_{ij} \) represents the number of pixels whose ground truth class is \( i \) and predicted class is \( j \), and \( C \) is the total number of classes.
\begin{figure*}[!h]
    \centering
    \includegraphics[width=\textwidth]{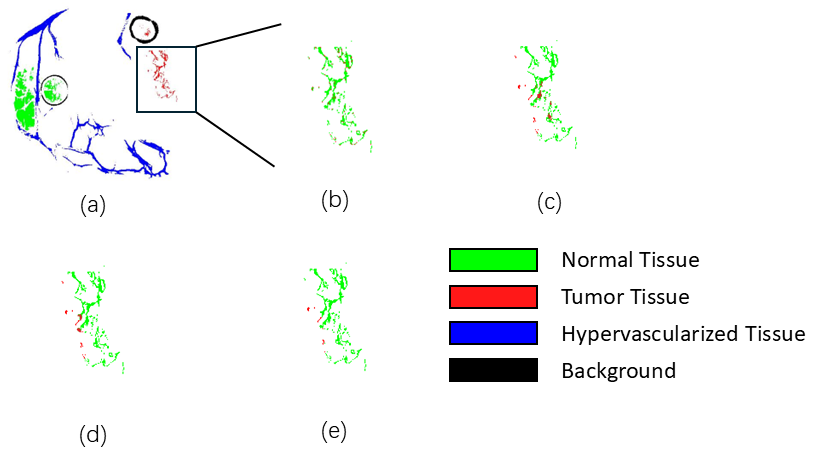} 
    %\caption{Classification results under different attacks and ground-truth on the brain dataset.(a) Ground-truth, (b) SS-FGSM, (c) SSA, (d) MfcaNet, (e) Ours.}
    \caption{Qualitative comparison of classification results on the Brain dataset under different adversarial attacks. (a) Ground truth. (b) SS-FGSM. (c) SSA. (d) MfcaNet. (e) Ours. The proposed method produces more pronounced classification shifts in lesion-related regions than competing attacks, while most non-lesion areas remain visually consistent, suggesting a more targeted disruption of lesion-sensitive spectral-spatial representations.}
    \label{fig:keshihua}
\end{figure*}

\textbf{Average Accuracy (AA)} calculates the mean classification accuracy across all classes, reflecting the model's balanced performance:
\begin{equation}
\text{AA} = \frac{1}{C} \sum_{i=1}^{C} \frac{N_{ii}}{\sum_{j=1}^{C}. N_{ij}}
\end{equation}
\textbf{Cohen's Kappa Score} is a statistical measure of agreement between predicted and true labels, adjusted for random chance:
\begin{equation}
\kappa = \frac{p_o - p_e}{1 - p_e},
\end{equation}
where the observed agreement \( p_o \) and expected agreement \( p_e \) are defined as:
\begin{equation}
p_o = \frac{\sum_{i=1}^{C} N_{ii}}{\sum_{i=1}^{C} \sum_{j=1}^{C} N_{ij}},
\end{equation}
\begin{equation}
p_e = \sum_{i=1}^{C} \left( \frac{\sum_{j=1}^{C} N_{ij}}{\sum_{i=1}^{C} \sum_{j=1}^{C} N_{ij}} \cdot \frac{\sum_{j=1}^{C} N_{ji}}{\sum_{i=1}^{C} \sum_{j=1}^{C} N_{ij}} \right).
\end{equation}

These metrics provide a comprehensive assessment of model performance, especially under adversarial conditions where lesion misclassification must be rigorously evaluated.

\subsection{Results Under Attacks and Defenses}
\label{sec:overall-results}
Unlike conventional adversarial attacks that aim to degrade overall model performance, our method focuses on inducing clinically critical errors. Our attack is designed with a targeted misclassification objective, where lesion regions (tumor/cancer) are intentionally misclassified as normal tissue, which represents a clinically critical error. Under this setting, a more significant reduction in the classification accuracy of tumor/cancer regions indicates a stronger attack performance. 

At the same time, we aim to preserve the correctness of non-lesion regions (e.g., normal tissue and other categories), resulting in high classification accuracy for these classes. However, it is important to note that our method does not explicitly constrain the predictions of all non-lesion categories. Instead, the attack objective is designed to prioritize the misclassification of lesion regions, without enforcing optimal performance on other classes. This selective behavior ensures that the perturbation remains structurally consistent and less perceptible, while focusing its impact on clinically critical regions rather than uniformly degrading all categories.

Due to this design, most non-lesion pixels are still correctly classified, leading to relatively high global metrics such as OA, AA, and Kappa. Notably, although the classification accuracy of certain non-lesion categories may not always achieve the best ranking, the differences compared to baseline methods are marginal. This indicates that our method does not cause a significant overall performance collapse, but instead introduces a targeted and localized degradation.

To provide a unified and mechanism-driven analysis, we jointly examine the results presented in Tab.~\ref{gu.t1} (Brain dataset) and Tab.~\ref{gu.t2} (MDC dataset). These two tables cover multiple classifiers (HybridSN, SSRN, SACNet, Dual-Stream) and representative defense models (RCCA, WFSS, AIAF, S\textsuperscript{3}ANet), enabling consistent observations across datasets and architectures. Discussing them together avoids fragmented reporting and highlights cross-cutting patterns that are obscured when each table is considered in isolation.  

Such a selective degradation is particularly meaningful in medical scenarios, where misclassifying lesion regions as normal may lead to severe clinical consequences, even when overall performance metrics remain high. A central finding is a lesion-first degradation pattern: our attack drastically reduces classification accuracy for lesion-related classes (tumor or cancer), while global metrics (OA/AA/Kappa) remain high due to class imbalance and smoothing effects of defenses. For example, under RCCA and WFSS, OA/AA/Kappa stay in the 96–99\% range, yet lesion accuracy drops most severely with our method (e.g., WFSS: $62.32\%$ tumor accuracy on Brain, $45.27\%$ cancer accuracy on MDC), indicating a selective but clinically critical shift toward false negatives.  

It is worth emphasizing the extreme values observed in Tables~\ref{gu.t1} and \ref{gu.t2}, where some non-lesion categories exhibit nearly perfect classification accuracy across baseline classifiers. This does not indicate a failure of the attack but reflects two intrinsic properties of the setting. First, standard HSI classifiers such as HybridSN, SSRN, SACNet, and Dual-Stream already achieve near-perfect accuracy on clean images for non-lesion tissues (Normal, Hypervascularized, Background in Brain; Normal in MDC). Since the targeted loss function predominantly drives the gradient updates towards lesion regions, the perturbations naturally concentrate on these areas, leaving non-lesion classes largely unaffected. In contrast, defense-oriented networks (RCCA, WFSS, AIAF, S\textsuperscript{3}ANet) cannot achieve strict $100\%$ accuracy on clean images due to the robustness–accuracy trade-off, and thus exhibit slight accuracy degradation after attacks. These patterns highlight the selectivity of our attack rather than any ineffectiveness.  

On the Brain dataset, tumor accuracy drops to single digits across all classifiers: $7.98\%$ for HybridSN, $4.65\%$ for SSRN, and $8.46\%$ for SACNet. The lowest value for SSRN suggests that residual spectral–spatial coupling is particularly vulnerable to locally coherent, multiscale perturbations, amplifying boundary shifts in lesion regions.  

On the MDC dataset (Normal vs.\ Cancer), our method maintains nearly $100\%$ accuracy for Normal while sharply degrading Cancer performance (e.g., $10.31\%$, $13.62\%$, $9.56\%$, and $32.64\%$ accuracy across classifiers). This constitutes a targeted shift from positive to negative—precisely the most harmful clinical error mode—mirroring the Brain results and aligning with the design intent of our attack.  

These behaviors are consistent with the attack design. The local pixel dependency component averages gradients within small neighborhoods, preserving anatomical coherence and visual plausibility. The multiscale component injects perturbations across multiple resolutions and reprojects them back, jointly shifting decision boundaries without introducing conspicuous artifacts. This resolves the paradox of high OA/AA/Kappa alongside catastrophic lesion-class collapse.  

Finally, qualitative evidence in Fig.~\ref{fig:keshihua} corroborates the quantitative findings: baseline methods produce only partial errors, whereas our approach induces extensive lesion misclassification while keeping least perturbations.  
\begin{figure*}[!h]
    \centering
    \includegraphics[width=\textwidth,height=0.3\textheight]{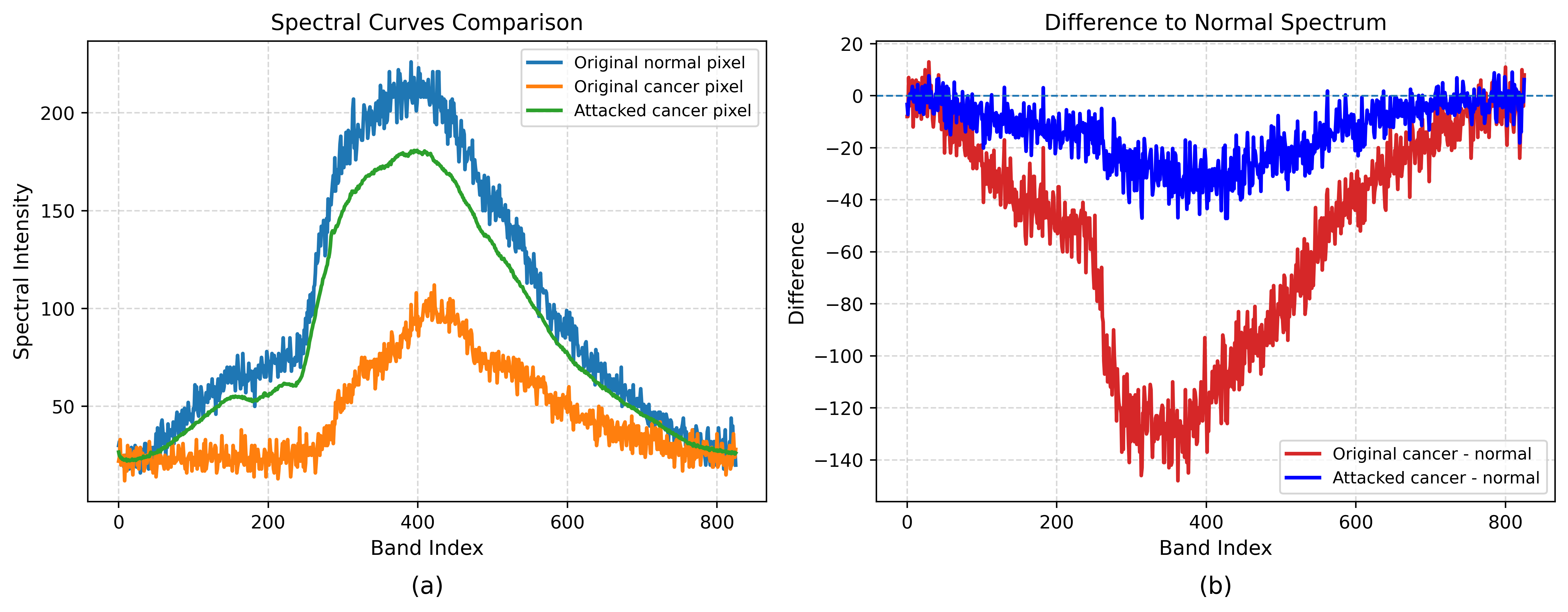} 
    %\caption{Spectral curves and difference analysis of normal, cancer, and adversarial cancer pixels.}
    \caption{Spectral analysis of normal, cancer, and adversarial cancer pixels. \textbf{Left:} The spectral signatures of a normal pixel, a cancer pixel, and the corresponding adversarially perturbed cancer pixel. After attack, the adversarial cancer spectrum shifts toward the normal spectral pattern while preserving a relatively smooth spectral profile. \textbf{Right:} The spectral differences with respect to the normal pixel, showing that the perturbation substantially reduces the discrepancy between cancer and normal spectra across most bands. This result suggests that the proposed attack induces lesion-to-normal misclassification by selectively aligning lesion spectra with normal tissue characteristics.}
    \label{fig:spectral_shift}
\end{figure*}

\subsection{Spectral Analysis of Adversarial Perturbations.}
To better understand how the proposed attack manipulates medical hyperspectral data, we further analyze the spectral characteristics of pixels before and after the attack. ~\figref{fig:spectral_shift} shows the spectral curves of a normal pixel, a cancer pixel, and the adversarially perturbed cancer pixel. 

As illustrated in the left panel, the spectral signature of the cancer pixel is significantly different from that of normal tissue across multiple spectral bands. After applying the proposed adversarial perturbation, the spectral curve of the cancer pixel shifts noticeably toward the normal spectral pattern while maintaining a smooth spectral structure. This indicates that the attack does not introduce random noise but instead generates structured perturbations that mimic the spectral characteristics of normal tissue.

The right panel presents the spectral differences relative to the normal pixel. The original cancer spectrum exhibits large deviations from the normal spectrum, whereas the adversarial cancer spectrum shows substantially reduced differences across most spectral bands. This shift in spectral characteristics explains why the classifier tends to misclassify lesion regions as normal tissue. The result further demonstrates that the proposed attack effectively exploits spectral–spatial dependencies to induce clinically critical misclassification while maintaining imperceptible perturbations.

\subsection{Ablation Study}
In this section, we conduct an ablation study to systematically evaluate both the effectiveness of the main components in our proposed adversarial attack framework and the sensitivity of its key design parameters. Specifically, Tab.~\ref{gu.t5} reports the component-level ablation results, including the effects of the spectral and spatial attention mechanisms as well as the Local Pixel Dependency Attack and the Multiscale Information Attack. In addition, Tab.~\ref{tab:scale_factor_ablation} and Tab.~\ref{tab:window_size_ablation} further analyze the influence of two important hyperparameters, namely the scale factors used in the multiscale attack and the neighborhood window size used in local gradient aggregation.
\subsubsection{Component Ablation}

To evaluate the contribution of different components in our proposed adversarial attack framework, we conduct a component level ablation study by selectively removing or combining the spectral–spatial attention mechanisms and the two attack modules. The results are summarized in Tab.~\ref{gu.t5}.

We first analyze the role of the attention mechanisms. When both spatial and spectral attention are removed, the tumor prediction accuracy remains relatively high at 30.67\%, indicating limited attack effectiveness. Removing only spectral attention further increases the tumor prediction accuracy to 26.49\%, demonstrating that spectral attention plays a critical role in exploiting spectral dependencies in hyperspectral data. When both spatial and spectral attention mechanisms are enabled (with the local attack only), the tumor prediction accuracy is reduced to 12.76\%, highlighting their complementary effects in guiding perturbation generation.

We then examine the contributions of the Local Pixel Dependency Attack and the Multiscale Information Attack. When only the multiscale attack is applied, the tumor prediction accuracy is reduced to 15.48\%, showing its effectiveness in disrupting hierarchical spectral–spatial features. The local pixel dependency attack alone achieves a tumor prediction accuracy of 12.76\%, indicating its strong capability in exploiting local spatial consistency. When both components are combined, the tumor prediction accuracy further drops to 8.46\%, demonstrating a clear synergistic effect between local structural modeling and multiscale perturbation.

It is also observed that removing attention mechanisms slightly degrades the classification accuracy of non-lesion tissues (e.g., Normal tissue drops from ~99\% to 85.94\%), indicating that the attention mechanism helps maintain the structural coherence of the entire image during perturbation.

Overall, the best performance is achieved when all components are jointly employed, yielding the lowest tumor prediction accuracy (8.46\%). This confirms that attention-guided perturbation, local dependency modeling, and multiscale representation disruption are all essential for maximizing the effectiveness of adversarial attacks in MHSI.

\begin{table*}[!ht]
\centering
\caption{Performance Comparison with Different Attention Mechanisms and Methods.}
\resizebox{\textwidth}{!}{%
\begin{tabular}{cccccccc}
\toprule
\textbf{\makecell{Spatial\\Attention}} & \textbf{\makecell{Spectral\\Attention}} & \textbf{\makecell{Local\\Pixel}} & \textbf{Multiscale} & \textbf{\makecell{Norma\\Tissue(↑)}} & \textbf{\makecell{Tumor\\Tissue(↓)}} & \textbf{\makecell{Hyper\\vascularized(↑)}} & \textbf{Background(↑)} \\ 
\midrule
\checkmark  & × & \checkmark  & \checkmark  & 85.94 & 30.67 & 84.24 & 89.76 \\ 
× & \checkmark  & \checkmark  & \checkmark  & 79.04 & 26.49 & 75.97 & 83.43 \\ 
\checkmark  & \checkmark  & \checkmark  & × & 95.31 & 15.48 & 91.74 & 94.98 \\ 
\checkmark  & \checkmark  & × & \checkmark  & 98.49 & 12.76 & 95.06 & 96.24 \\ 
\checkmark  & \checkmark  & \checkmark  & \checkmark  & 94.89 & \textbf{8.46} & 89.25 & 94.51 \\ 
\bottomrule
\end{tabular}%
}

\label{gu.t5}
\end{table*}
\subsubsection{Sensitivity Analysis of Scale Factors and Window Size}

To investigate the impact of key hyperparameters in our adversarial attack framework, we conduct a sensitivity analysis on (1) the set of scale factors \( S \) used in the Multiscale Information Attack and (2) the window size \( N \) employed in the Local Pixel Dependency Attack. These parameters respectively control the diversity of multiscale perturbations and the spatial extent of local structural modeling.

\textbf{(a) Effect of Scale Factors \( S \):}

We evaluate the framework using four different scale sets:
\[
S_1 = \{1\},\quad S_2 = \{1, 2\},\quad S_3 = \{1, 2, 4\},\quad S_4 = \{1, 2, 4, 8\}.
\]

As shown in Tab.~\ref{tab:scale_factor_ablation}, incorporating multiple scales significantly improves the effectiveness of the attack. Using only a single scale (\( S = \{1\} \)) results in limited performance (50.87\%), as the perturbation is restricted to a single resolution and fails to capture hierarchical spectral–spatial patterns. Expanding the scale set to \( \{1,2\} \) and \( \{1,2,4\} \) progressively enhances the attack strength, demonstrating that multiscale perturbations are crucial for disrupting features at different resolutions.

However, when introducing excessively large scales (\( S = \{1,2,4,8\} \)), the performance degrades (16.36\%). This is because overly coarse scales tend to introduce globally smooth perturbations, which dilute fine-grained structural details and reduce the attack's ability to precisely target lesion regions. Therefore, \( S = \{1,2,4\} \) achieves the best balance between multiscale diversity and perturbation precision, and is adopted in our final configuration.

\textbf{(b) Effect of Window Size \( N \):}

To assess the sensitivity to local spatial context, we vary the window size \( N \) in the Local Pixel Dependency Attack as:
\[
N = 3 \times 3,\ 5 \times 5,\ 7 \times 7,\ 9 \times 9,\ 11 \times 11,\ 13 \times 13,\ 15 \times 15.
\]

As shown in Tab.~\ref{tab:window_size_ablation}, the window size has a significant impact on the effectiveness of local structural modeling. Small windows (e.g., \( 3 \times 3 \)) provide limited spatial context and fail to capture sufficient local dependencies, resulting in weaker attack performance. As the window size increases, the attack becomes more effective, indicating that incorporating richer local neighborhood information helps preserve structural consistency while guiding more coherent perturbations.

However, excessively large windows (e.g., \( 13 \times 13 \) and \( 15 \times 15 \)) lead to performance degradation. This is because overly large neighborhoods introduce excessive smoothing in the gradient aggregation process, which weakens local discriminative patterns and reduces the precision of perturbations. As a result, a moderate window size such as \( 11 \times 11 \) provides the best trade-off between structural coherence and attack sharpness, aligning with the intrinsic locality of anatomical structures in MHSI.

\begin{table}[ht]
\centering
\caption{Effect of Scale Factors \( S \) on Attack Effectiveness (Tumor Class Accuracy ↓)}
\label{tab:scale_factor_ablation}
\begin{tabular}{c|c}
\hline
Scale Factors \( S \)  & Tumor Acc. (\%) (↓)  \\
\hline
\{1\}    &  50.87    \\
\{1, 2\}    & 27.43     \\
\{1, 2, 4\}    &{\textbf{9.27}}      \\
\{1, 2, 4, 8\}    & 16.36     \\
\hline
\end{tabular}
\end{table}

\begin{table}[ht]
\centering
\caption{Effect of Window Size \( N \) on Attack Effectiveness (Tumor Class Accuracy ↓)}
\label{tab:window_size_ablation}
\begin{tabular}{c|c}
\hline
Window Size \( N \)  & Tumor Acc. (\%) (↓)  \\
\hline
$3 \times 3$   &  31.62   \\
$5 \times 5$   &  22.48   \\
$7 \times 7$   &  17.36   \\ 
$9 \times 9$   &  14.21   \\
$11 \times 11$ &  {\textbf{10.87}}   \\
$13 \times 13$ &  13.74   \\
$15 \times 15$ &  18.92   \\
\hline
\end{tabular}
\end{table}

\section{Conclusion}
This study introduces a specialized adversarial attack framework specifically designed for MHSI, addressing the unique spectral-spatial characteristics and multiscale features inherent in medical data. Our innovative Local Pixel Dependency Attack leverages precise spatial relationships between neighboring pixels, while the Multiscale Information Attack strategically targets hierarchical spectral-spatial features. These innovations effectively exploit critical vulnerabilities in medical deep learning classifiers, significantly reducing classification accuracy for clinically relevant tumor regions on Brain and MDC datasets, outperforming existing methods such as SS-FGSM, SSA, and MfcaNet. However, our approach could be further enhanced by incorporating domain-specific priors, such as spectral similarity between tumor and surrounding tissues, to refine perturbation precision. In addition, under the small perturbation budget, together with the structured perturbation design, the generated adversarial examples preserve relatively smooth and coherent spectral patterns rather than exhibiting abrupt spectral distortions. Nevertheless, our current framework does not explicitly enforce physical reflectance constraints, future work could integrate physics-informed regularization terms into the perturbation generation process to ensure strict spectral plausibility.

The clinical relevance of our method is substantial, as adversarial misclassifications of tumor regions can critically affect diagnostic accuracy, leading to potential misdiagnoses and compromised patient outcomes. By explicitly addressing vulnerabilities related to spectral-spatial dependencies and multiscale information, this research highlights the urgent need for robust defensive strategies tailored specifically to  MHSI-based diagnostic systems. Future research should further validate the clinical applicability of our method in diverse scenarios and develop targeted defensive measures to enhance the reliability and safety of medical imaging practices.

% acknowledgments part

\section{Abbreviations}
OA: Overall Accuracy; AA: Average Accuracy; KAPPA: Cohen’s Kappa Score; CNN(s): Convolutional Neural Network(s); DNNs: Deep Neural Networks; HSI: Hyperspectral Image; KNN: k-Nearest Neighbors; MDC: Multidimensional Choledoch; MHSI: Medical Hyperspectral Imaging; MLP: Multilayer Perceptron;  PCA: Principal Component Analysis; SAM: Spectral Angle Mapper; SSFCN: Spectral-Spatial Fully Convolutional Networks; SSRN: Spectral-Spatial Residual Network; SVM: Support Vector Machines.

\section{Declarations}
\subsection{Data Availability}
The datasets analyzed during the current study are available in the MHSI Choledoch Dataset and In-vivo HS Human Brain database, [https://www.kaggle.com/datasets/hfutybx/mhsi-choledoch-dataset-preprocessed-dataset/data, \\https://hsibraindatabase.iuma.ulpgc.es/].

\subsection{Competing Interests}
The authors have no relevant financial or non-financial interests to disclose.

\subsection{Author Contributions}
All authors contributed to the study conception and design. Material preparation, data collection and analysis were performed by Yunrui Gu. The first draft of the manuscript was written by Yunrui Gu, Zhenzhe Gao and Cong Kong. All authors commented on previous versions of the manuscript. All authors read and approved the final manuscript.

\subsection{Funding}
This work was supported by the National Natural Science Foundation of China, No. 62472177, and the Shanghai Sci-tech Co-research Program, No. 251222020A.
\begin{acknowledgements}
Not applicable.
\end{acknowledgements}

% BibTeX from reference.bib
%\bibliographystyle{sn-apacite}
\bibliographystyle{unsrt}
\bibliography{reference}

\end{document}